\documentclass[lettersize,journal]{IEEEtran}
\usepackage{amsmath,amsfonts}
\usepackage{algorithmic}
\usepackage{algorithm}
\usepackage{array}
\usepackage{textcomp}
\usepackage{stfloats}
\usepackage{url}
\usepackage{verbatim}
\usepackage{graphicx}
\usepackage{cite}
\usepackage{bbm}
\usepackage{subcaption}
\newcommand{\R}{\mathbbm{R}}
\newcommand{\bX}{\mathbf{X}}
\newcommand{\bx}{\mathbf{x}}

\begin{document}

\title{Tukey g-and-h neural network regression for non-Gaussian data}

\author{A. P. Guillaumin, N. Efremova
\thanks{}
\thanks{}}



\maketitle

\begin{abstract}
This paper addresses non-Gaussian regression with neural networks via the
use of the Tukey g-and-h distribution. 
The Tukey g-and-h transform is a flexible
parametric transform with two parameters $g$ and $h$ which, when applied to a standard normal random
variable, introduces both skewness and kurtosis, resulting
in a distribution commonly called the Tukey g-and-h distribution.
Specific values of $g$ and $h$
produce good approximations to other families of distributions, such as the Cauchy and student-t distributions.
%
The flexibility of the Tukey g-and-h distribution has 
driven its popularity in the statistical community, 
in applied sciences and finance. In this work we consider
the training of a neural network to predict the
parameters of a Tukey g-and-h distribution in a regression
framework via the minimization of the corresponding 
negative log-likelihood,
despite the latter having no closed-form expression.
We demonstrate the efficiency of our procedure in simulated examples and apply our method
to a real-world dataset of global crop yield for several types of
crops.
Finally, we show how we can carry out a goodness-of-fit analysis
between the predicted distributions and the test data.
A Pytorch implementation is made available
on Github and as a Pypi package.
\end{abstract}


\section{Introduction}
\IEEEPARstart{T}{he} application of Deep Learning to highly
non-linear regression tasks has brought
unprecedented results in many real-world
applications. 
In many practical regression problems, it is desirable to predict
a probabilistic distribution rather than a single-point value. This is key 
when machine learning algorithms are used in complex decision-making 
processes. 
For instance, if a neural network is used to
predict a stock price or
the duration of a surgical operation
\cite{pmlr-v68-ng17a}, confidence
intervals will be far more valuable than
single point predictions.
One must distinguish between
different sources of uncertainty:
model estimation uncertainty
--- which we do not consider as it is not the main topic
of this paper but which
 could be addressed via Bayesian
Neural Networks~\cite{bayesiannn} or Bootstrapping~\cite{NEURIPS2020_492114f6}
---
 and what one might call inherent uncertainty, that is, 
the remaining randomness even if the true regression 
model was known exactly.
We note that inherent uncertainty may not only arise
from the generating process, but also from the potentially
limited set of features used for regression.
One common approach to account for inherent uncertainty
consists in predicting, for
a given set of features, the two parameters of a Gaussian distribution, rather than a single-point value.
This is particularly valuable in the case where the
conditional variance of the target also depends on the features.
However, the Gaussian assumption itself might be quite restrictive, as
it does not allow for skewness and kurtosis of the target
variable
conditioned on the features. 
These patterns may arise in the target variable either
from the generating process itself, or from using
an incomplete set of features.
One way to address non-Gaussianity in the 
target variable is to consider
more flexible parametric families of distributions. 
For instance, one might train a neural network
to output the parameters of a mixture of Gaussian distributions
rather than a single Gaussian distribution.
This might be useful in particular when the conditional
distribution of the target is multi-modal, but
remains limited, in particular when it comes 
to modelling heavy tails. It is also worth
mentioning non-parametric approaches such
as quantile regression~\cite{XU2017129, quantilereg}.
In this paper, however, we focus on the parametric
approach, more specifically on the Tukey g-and-h probability
distribution~\cite{gandhproperties}.
This distribution
is obtained by applying the Tukey g-and-h transform with parameters
$g$ and $h$ to a standard normal random variable, which is then 
rescaled by a multiplicative factor $\sigma$ and shifted by an additive constant $\mu$. As such, the Tukey g-and-h distribution
has four parameters. 
The Tukey g-and-h distribution has been widely used
in modern statistical 
geosciences~\cite{xu2017tukey, jeong2019stochastic}, spatio-temporal
modelling~\cite{9153778}, but also in financial
risk analysis~\cite{riskanalysis}.
The standard random field 
approach~\cite{xu2017tukey} consists in modelling spatio-temporal
dependence via a Gaussian Process, followed by a pointwise
mapping by the Tukey g-and-h transform to incorporate non-Gaussian
patterns.
In parallel,
the application of neural networks to 
geosciences has grown exponentially, for instance
to address problems such as hyper-resolution~\cite{superresolution1},
forecasting~\cite{forecast1} or for the parameterization
of discretized non-linear PDE 
solvers~\cite{parameterization1} and yield forecasting \cite{you2017deep}, \cite{dado2020high}, \cite{wang2020mapping}.
The objective of this work is to propose a method 
that bridges the gap between Tukey g-and-h random
fields and neural network regression. In particular,
we see a longer-term incentive to combine Neural Networks to learn
complex feature-dependent parameters of a Tukey g-and-h transform
with Gaussian Process models to incorporate residual patterns
of spatio-temporal dependence.
Our paper is organized as follows.
In Section 2, we review the Tukey g-and-h transform and
present our methodology for training and evaluating neural networks for the prediction of g-and-h 
distributions. 
This requires us to obtain the derivatives of the
Tukey g-and-h log likelihood with respect to the parameters,
which involves the inversion of the Tukey g-and-h transform.
There is no known closed form for the latter.
\IEEEpubidadjcol
We propose the use
of binary search to address this issue, which is
efficient both numerically and computationally.
In Section 3, we demonstrate the benefits of the 
proposed methodology based on simulated data experiments,
while in Section 4 we present an application to a real-world
regression problem by applying our methodology 
to learn patterns of global yield for several crops.
Finally, we provide a Pytorch implementation
of the Tukey g-and-h loss function
made publicly available on the Github
account of the authors.

\section{Methodology}
In this section, we present our proposed methodology for the use of the
Tukey g-and-h transform in Neural Network
regression. We first review the
Tukey g-and-h transform and its basic properties.
We then consider
the evaluation of the negative log-likelihood function, which
we will use as our loss function for training. 
In evaluating the negative log-likelihood, the main challenge
lies
in the inversion of the Tukey g-and-h transform. We propose to use  
binary search
to address this problem. In comparison to other approaches proposed to 
approximate the inverse
of the Tukey g-and-h transform
such as grid
search, this entails no approximation other than that incurred by numerical
precision.
Finally, we discuss how one can assess goodness-of-fit and obtain
confidence intervals in Tukey g-and-h Neural Network regression.

\subsection{The Tukey g-and-h transform}
We review some properties of 
the Tukey g-and-h transform~\cite{xu2017tukey, gandhproperties}
and its ability to approximate a wide
range of other well-known families of 
probability distributions, when applied
to a standard normal random variable.

The Tukey g-and-h
transform with parameters $g\in\R$ and $h\geq 0$ is defined by,
\begin{equation}
    \tau_{g, h}(z) = 
    \frac{\exp(gz) - 1}{g}
    \exp\left(\frac{1}{2}hz^2\right),
    \quad
    \forall z\in\R,
\end{equation}
when $g \neq 0$. The case $g=0$ is obtained
by continuous extension,
\begin{equation*}
    \tau_{0, h}(z)
    =
    z
    \exp\left(\frac{1}{2}hz^2\right),
    \quad
    \forall z\in\R.
\end{equation*}
When both $g$ and $h$ are zero, the
transform is just the identity function.
For fixed $g$ and  $h$ the transform
$\tau_{g,h}(z)$ is an increasing function in $z$.

\begin{figure}
    \centering
    \subfloat[]{
        \includegraphics[width=0.35\textwidth]{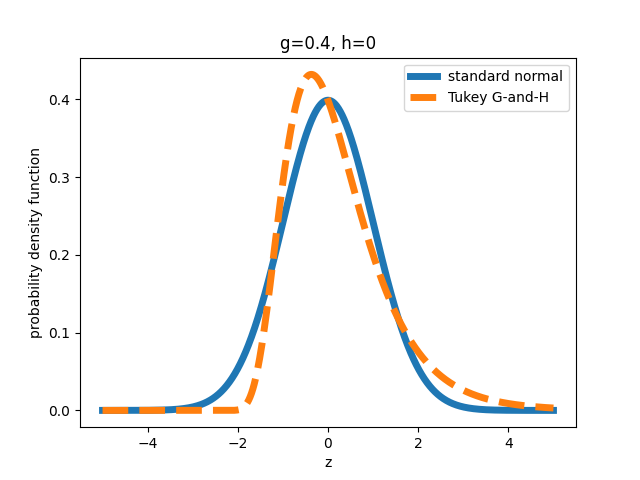}
        \label{fig:pdfa}
    }\\
    \subfloat[]{
        \includegraphics[width=0.35\textwidth]{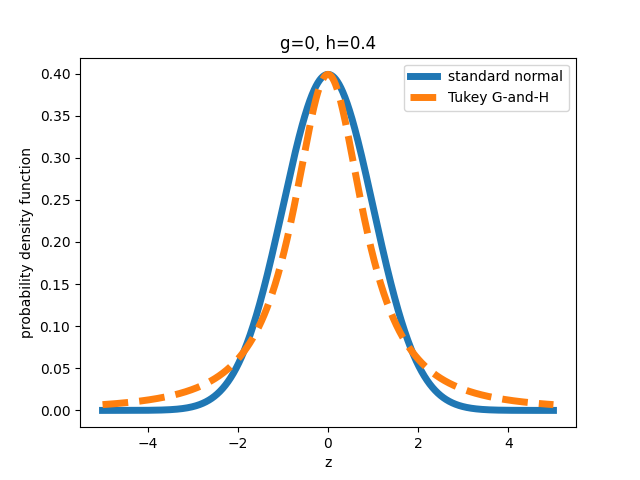}
        \label{fig:pdfb}
    }\\
    \subfloat[]{
        \includegraphics[width=0.35\textwidth]{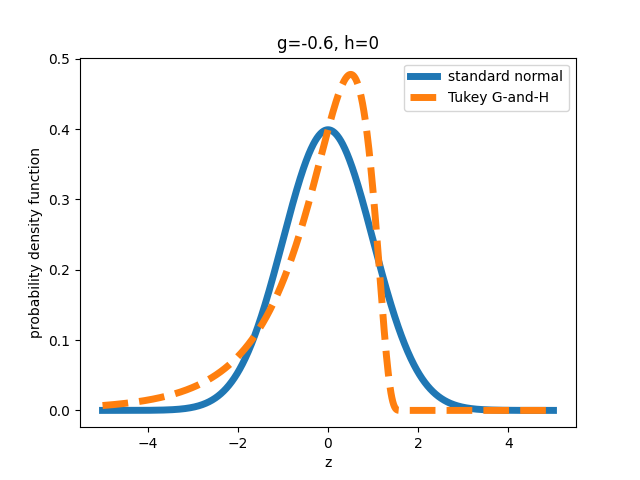}
        \label{fig:pdfc}
    }\\
    \subfloat[]{
        \includegraphics[width=0.35\textwidth]{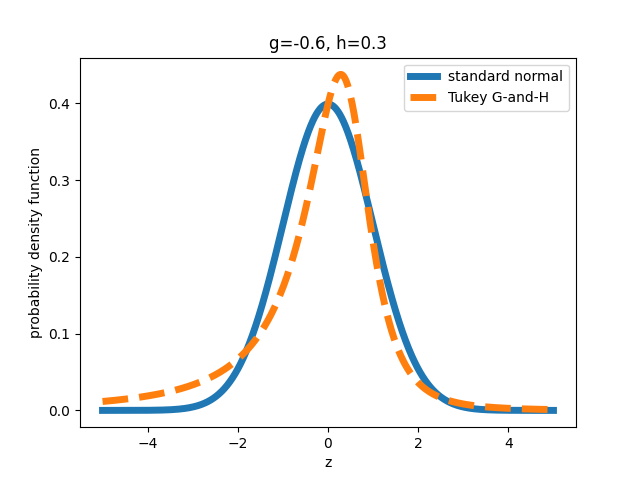}
        \label{fig:pdfd}
    }
    \caption{Comparison of the probability
    density function of a Tukey g-and-h distributed random
    variable with that of a standard normal random
    variable, for different values of $g$ and $h$.}
    \label{fig:pdfs}
\end{figure}

Let $Z$ be a standard normal random variable, and define,
\begin{equation}
\label{eq:defY}
\begin{cases}
    \widetilde{Z} &= \tau_{g, h}(Z),\\
    Y &= \mu + \sigma \widetilde{Z}.
\end{cases}
\end{equation}
where $\mu\in\R$ and $\sigma\in\R$ roughly control the first two moments of the transformed random
variable $Y$. The parameters $g$ and $h$ roughly control, respectively, the
skewness and kurtosis, which are zero for the non-transformed
variable $Z$. 
A positive value of $g$ will incur positive skewness in 
$Y$ --- see Figure~\ref{fig:pdfa}--- while a negative value of
$g$ will incur negative skewness  --- see Figure~\ref{fig:pdfc}.
Note that as $g$ and $h$ converge to zero, $Y$
converges in distribution towards a Gaussian random variable with 
mean $\mu$ and standard deviation $\sigma$.
In this manuscript, we will say that the transformed random 
variable $\widetilde{Y}$ follows a G-and-H distribution.
In Figure~\ref{fig:pdfs} we show the probability density function of 
$\widetilde{Z}$ for 4 combinations of the 
parameters $g$ and $h$, superimposed
with the probability density function
of a standard normal distribution.
The Tukey g-and-h distribution provides a
good approximation to a wide range of standard 
probability distributions~\cite{gandhproperties}.

\subsection{Training via negative likelihood loss
minimization}
\label{sec:lkh}
We consider a regression problem with features $\bX$ and 
target $Y$. Rather than only training a neural network
to predict the conditional expectation of $Y$ given
$\bX$, we wish to approximate the full conditional distribution
of $Y$ given $\bX$ by training the neural network
to predict a g-and-h distribution --- that is, predict
in terms of $\bX$
the four parameters 
$\mu, \sigma, g$ and $h$
of such a distribution. As such, the output
of our neural network should have a total of $4$ neurons,
one for each parameter of a g-and-h distribution.
In this section, we provide details on how to train
a neural network for Tukey g-and-h regression
via stochastic gradient
descent on the negative log-likelihood.
Following standard properties of transformed random
variables, the transformed random variable $Y$ 
defined in~\eqref{eq:defY}
admits a probability density
function which can be expressed in terms
of the probability density
function 
$f_Z(z) = \frac{1}{\sqrt{2\pi}}\exp(-z^2/2)$
of the standard normal random variable $Z$.
Specifically,
\begin{equation*}
    f_{Y}(y)
    =
    \frac{1}{
        \sigma\ \tau_{g, h}'
        \left(
            \tau_{g, h}^{-1}
            \left(
                \frac{y - \mu}{\sigma}
            \right)
        \right)
    }
    f_Z
    \left(
        \tau_{g, h}^{-1}
        \left(
            \frac{y - \mu}{\sigma}
        \right)
    \right),
\end{equation*}
where $\tau_{g, h}'(z)$ is the derivative of 
the Tukey g-and-h transform $\tau_{g, h}(\cdot)$
given by,
\begin{equation*}
    \tau_{g, h}'(z) = 
    \left[
        \exp(gz) + hz\frac{\exp(gz) - 1}{g}
    \right]
    \exp\left(
        \frac{1}{2}hz^2
    \right),
\end{equation*}
and $\tau_{g, h}^{-1}$ is the inverse function of the
Tukey g-and-h transform for fixed parameters $g$ and $h$. 

Given a collection of features $\mathbf{x}_i$ and response variables $y_i$,
$i=1,\ldots,n$, we denote $\mu_i, \sigma_i, 
g_i$ and $h_i$ the 4 components of the neural 
network's output for input $\bx_i$, corresponding
to the four parameters of a G-and-H distribution.
If we denote
$\widehat{z}_i=
\tau_{g_i, h_i}^{-1}
\left(
    \frac{y - \mu_i}{\sigma_i}
\right),
$
the negative log-likelihood function can 
be expressed in terms of
the neural network's parameters $\theta$ as,
\begin{align}
    \nonumber
    l(\theta) &= 
    \sum_i
    \log \tau_{g_i, h_i}'(\hat{z}_i)
    +
    \sum_i\log \sigma_i
    +\frac{1}{2}\sum_i\hat{z}_i^2\\
    &=
    \label{eq:loglkh}
    \nonumber
    \sum_i
    \log\left[
        \exp(g\hat{z}_i)
        + h\hat{z}_i\frac{\exp(g\hat{z}_i) - 1}{g}
    \right]
    \\
    &+
    \sum_i\log \sigma_i
    +\sum_i
    \frac{1 + h_i}{2}\hat{z}_i^2.
\end{align}
The dependence of $g_i$, $h_i$, $\sigma_i$ and $\hat{z}_i$ 
on the neural
network's parameters $\theta$ is
left implicit in the notation for simplicity.
The evaluation of the $\widehat{z}_i$'s
and their gradient with respect to the parameters
$g$ and $h$
 ---
these are required to carry out Stochastic 
Gradient Descent ---
poses the main challenge
to the evaluation of the loss. 
One previously proposed approach is to approximate the inverse 
Tukey g-and-h transform using a grid of values $z_0, \ldots, z_p$ for which $\tau_{g,h}$
is evaluated~\cite{xu2017tukey}.
In contrast, we propose an approach which is
not limited in accuracy other than by that of the precision of the numerical implementation.
More specifically, since the Tukey g-and-h transform is an increasing function of $z$, its inverse at any given
point can be obtained efficiently by binary search. 
The only requirement is that we start the algorithm with a large enough range.

Finally, in order to apply Stochastic Gradient Descent algorithms,
we also require the derivatives of the inverse transform with 
respect to $g$, $h$ and $\widetilde{z}$
due to the terms $\hat{z}_i$
in the loss function~\eqref{eq:loglkh}.
First, we have,
\begin{equation}
    \frac
    {\partial\tau_{g,h}^{-1}}
    {\partial\widetilde{z}}
    (\widetilde{z})
    =
    \frac{1}{
        \tau_{g,h}'(\tau_{g,h}^{-1}(\widetilde{z}))
    }.
\end{equation}
Then, we derive,
\begin{align}
    \frac{\partial \tau_{g, h}}{\partial g}(z) 
    &= \frac{\exp(gz)(gz - 1) + 1}{g^2}
    \exp(\frac{1}{2}hz^2),\\
    \frac{\partial \tau_{g, h}}{\partial h}(z)
    &= 
    \frac{z^2}{2}\tau_{g,h}(z).
\end{align}
We then write,
\begin{equation}
    \tau_{g, h}
    \left(
        \tau_{g, h}^{-1}(\widetilde{z})
    \right)
    =
    \widetilde{z},
\end{equation} 
and obtain the desired quantities after taking the derivative
of the above equation with
respect to $g$ and $h$ respectively, by application of
the chain rule,
\begin{align}
    \frac{\partial \tau_{g, h}^{-1}}{\partial g}(\widetilde{z})
    &=
    -
    \frac{
        \frac{\partial \tau_{g, h}}{\partial g}
        \left(
            \tau_{g,h}^{-1}(\widetilde{z})
        \right)
    }
    {
    \tau_{g,h}'(\tau_{g,h}^{-1}(\widetilde{z}))
    },\\
    \frac{\partial \tau_{g, h}^{-1}}{\partial h}(\widetilde{z})
    &=
    -
    \frac{
        \frac{\partial \tau_{g, h}}{\partial h}
        \left(
            \tau_{g,h}^{-1}(\widetilde{z})
        \right)
    }
    {
        \tau'_{g,h}(\tau_{g,h}^{-1}(\widetilde{z}))
    }.
\end{align}
By defining a \emph{Function} object in PyTorch that
encapsulates the inverse Tukey transform and its
derivatives with respect to $g$, $h$ and $\widetilde{z}$, 
we can obtain the gradient of the negative
log-likelihood function via automatic differentiation.
The code provided alongside this manuscript
takes care of implementing these computations, so that the user
only needs to declare a TukeyGandHLoss
object
and provide the four parameters output by their
neural network for each data point of a mini-batch.

\subsection{Prediction intervals}
One benefit of the Tukey g-and-h distribution 
for the modelling of non-Gaussian
random variables is that we can easily obtain prediction intervals for the target
variable. Let $0<\alpha<1$. Due to the continuous and increasing
nature of the Tukey g-and-h transform,
we immediately obtain, 
\begin{equation}
    F_{\widetilde{Z}}^{-1}(\alpha) = \tau_{g,h}(\Phi^{-1}(\alpha)),
\end{equation}
where $F_{\widetilde{Z}}^{-1}$ is the inverse cumulative distribution function of
the transformed random variable, and $\Phi^{-1}$ is the inverse cumulative distribution
function of the standard normal distribution.

Denote $\widehat{\theta}$
the parameters of the trained neural network.
For an input $\bx$, write
\begin{equation}
    \mu(\bx;\widehat{\theta}), 
\sigma(\bx;\widehat{\theta}),
g(\bx;\widehat{\theta}) \text{ and }
h(\bx;\widehat{\theta})
\end{equation}
for the four parameters of the g-and-h distribution
output by the neural network with parameters
$\widehat{\theta}$ when provided with input $\mathbf{x}$.
An $\alpha$-level confidence interval for a feature
$\bx$ is provided by,
\begin{align*}
\left[
    \mu(\bx;\widehat{\theta})
    +
    \sigma(\bx;\widehat{\theta})
    \tau_{g(\bx;\widehat{\theta}),
    h(\bx;\widehat{\theta})}
    (
        - z_{1-\alpha/2}
    )\right.
, \\
    \left.
    \mu(\bx;\widehat{\theta})
    +
    \sigma(\bx;\widehat{\theta})
    \tau_{g(\bx;\widehat{\theta}),
    h(\bx;\widehat{\theta})}
    (
        z_{1-\alpha/2}
    )
\right],
\end{align*}
with $z_{\alpha}\equiv F_Z^{-1}(\alpha)$.
While this approach is simple and computationally efficient,
it was noted in~\cite{xu2017tukey} that it may lead to
unreasonably large prediction intervals when the skewness 
parameter $g$ is large in absolute value. Following their approach,
one might instead consider the following prediction interval
\begin{align*}
\left[
    \mu(\bx;\widehat{\theta})
    +
    \sigma(\bx;\widehat{\theta})
    \tau_{g(\bx;\widehat{\theta}),
    h(\bx;\widehat{\theta})}
    (
        - z_{1 -\alpha + \gamma}
    )\right.
, \\
    \left.
    \mu(\bx;\widehat{\theta})
    +
    \sigma(\bx;\widehat{\theta})
    \tau_{g(\bx;\widehat{\theta}),
    h(\bx;\widehat{\theta})}
    (
        z_{1-\gamma}
    )
\right],
\end{align*}
where $0\leq\gamma\leq\alpha$ is chosen so that it minimizes the length
of the prediction interval.
We note that the prediction intervals
obtained in this fashion 
do not account for the uncertainty
in $\widehat{\theta}$. 
While this is not the focus of this paper, a potential means of 
addressing this issue might be the use of 
bootstrapping~\cite{sluijterman2023confident}.

\subsection{Goodness-of-fit analysis}
\label{sec:gof}
The standard approach to goodness-of-fit analysis in Machine Learning relies on evaluating the 
loss function on a test dataset.
In the case where we predict a parametric probability distribution
rather than point values, we might also be interested in comparing the distribution of the 
predicted target values with that of the observed targets. In general, this poses some
difficulty as the predicted distribution of the target depends on the features.
In the case of Tukey g-and-h Neural Network regression,
however, we can easily address this problem.
We compare the empirical distribution of the $\hat{z}_i$'s to
the standard normal distribution
via standard methods such as quantile-quantile plots. 
Naturally, one should keep in mind 
that there could be compensation
effects in such an analysis: a
misfit could compensate another
misfit. 
We investigate this approach both
in the simulated and real-world data experiments of the next sections.
A somewhat equivalent approach is to compute 
\emph{uniform residuals} by applying the PIT 
(Probability Integral Transform) on the 
$\hat{z}_i$'s. Specifically, we define
$u_i = \Phi(\hat{z}_i)$ where $\Phi$
is used to denote the cumulative distribution
function of a standard normal random variable.
We can then compare the empirical distribution of the 
$u_i$'s to that of a uniform random variable on
the interval $[0, 1]$. Alternatively, we might
want to plot the $u_i$'s in terms of the features
--- ideally, we should not observe any pattern of dependence.

\section{Simulated data experiments}
In this section we present some results on Tukey g-and-h
neural network regression in a simulated setting.
We first study the case where the true 
distribution of the target variable, 
conditioned on the features, is indeed a
Tukey g-and-h distribution. We then assess the
robustness of the proposed methodology in a
misspecified setting where the 
target variable follows a student-t distribution, conditionally on the 
features.

\subsection{No model misspecification}
We first treat the case where the true
conditional distribution
of the target variable $Y$ actually belongs to the 
family of Tukey g-and-h distributions.
Specifically, we consider a scalar feature $X$
uniformly distributed on the interval $[0, 1]$, 
\begin{equation}
    X\sim \mathcal{U}([0, 1]).
\end{equation}
The response variable $Y$ conditionally on
$X$ is written as,
\begin{equation}
    \label{eq: ycondx}
    Y|X \sim \sigma(X) \tau_{g(X), h(X)}(Z) + \mu(X),
\end{equation}
with $Z\sim\mathcal{N}(0, 1)$. 
For the purposes of this simulated data experiment,
the deterministic
functions $\sigma(\cdot)$, $\mu(\cdot)$, $g(\cdot)$
and $h(\cdot)$ are set to arbitrarily chosen closed-form
functions. From these definitions, we simulate a dataset of size 40,000.

We define a neural network with 5 fully-connected layers,
with a scalar input (the feature $x$) and 4 outputs
corresponding to the 4 functions of $X$ 
that appear in~\eqref{eq: ycondx}.
We train this neural network using
training and validation data according to a standard
$80\%$ versus $20\%$ split. Exact details of the 
training procedure and functional form of the true regression
functions are available in the code provided as supplementary material.

\begin{figure}[h!]
    \centering
    \includegraphics[width=0.48\textwidth]{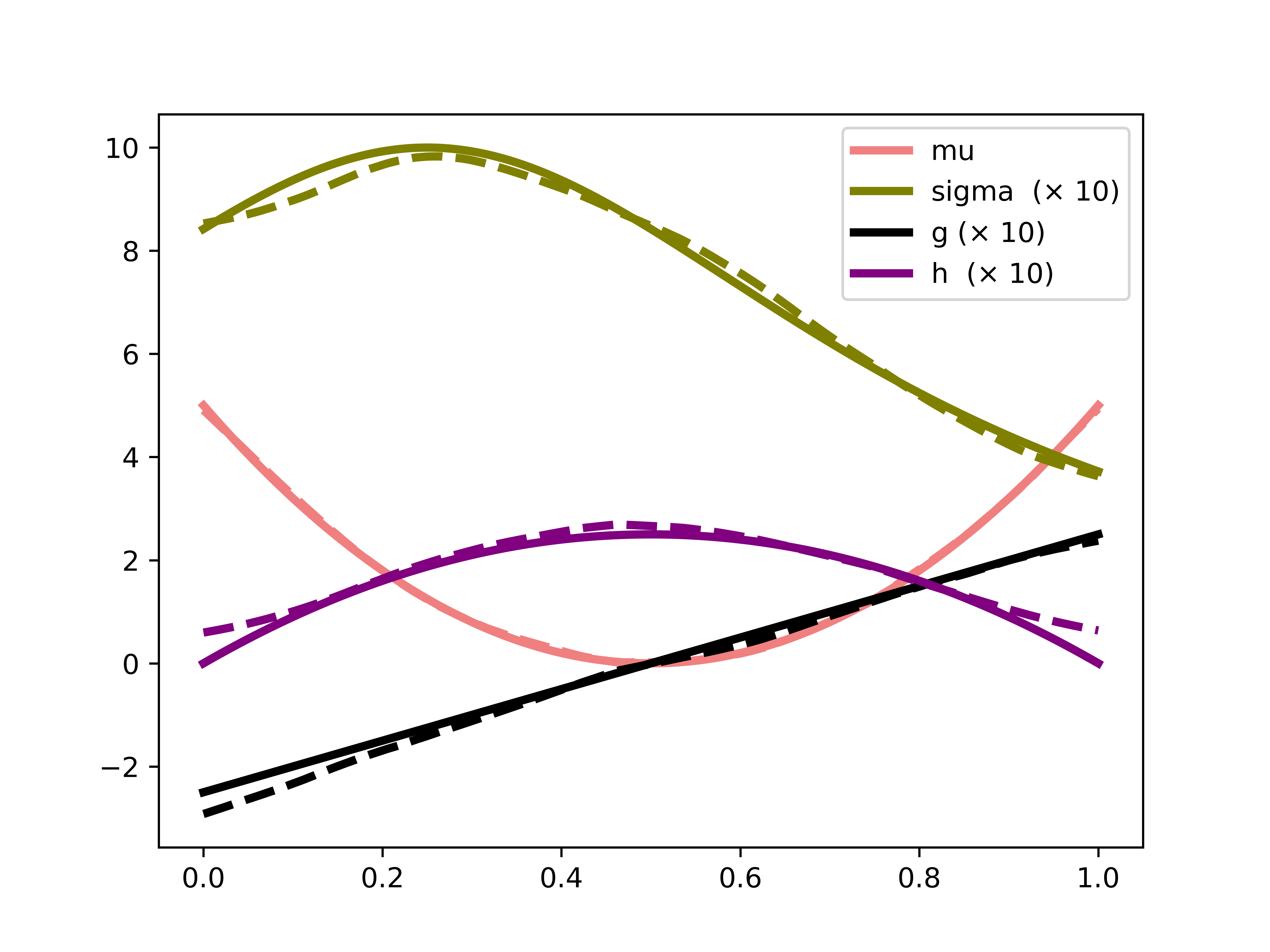}
    \caption{True regression functions (solid) and trained
    regression functions (dashed) on 10000 observations.}
    \label{fig:fig1}
    \includegraphics[width=0.48\textwidth]{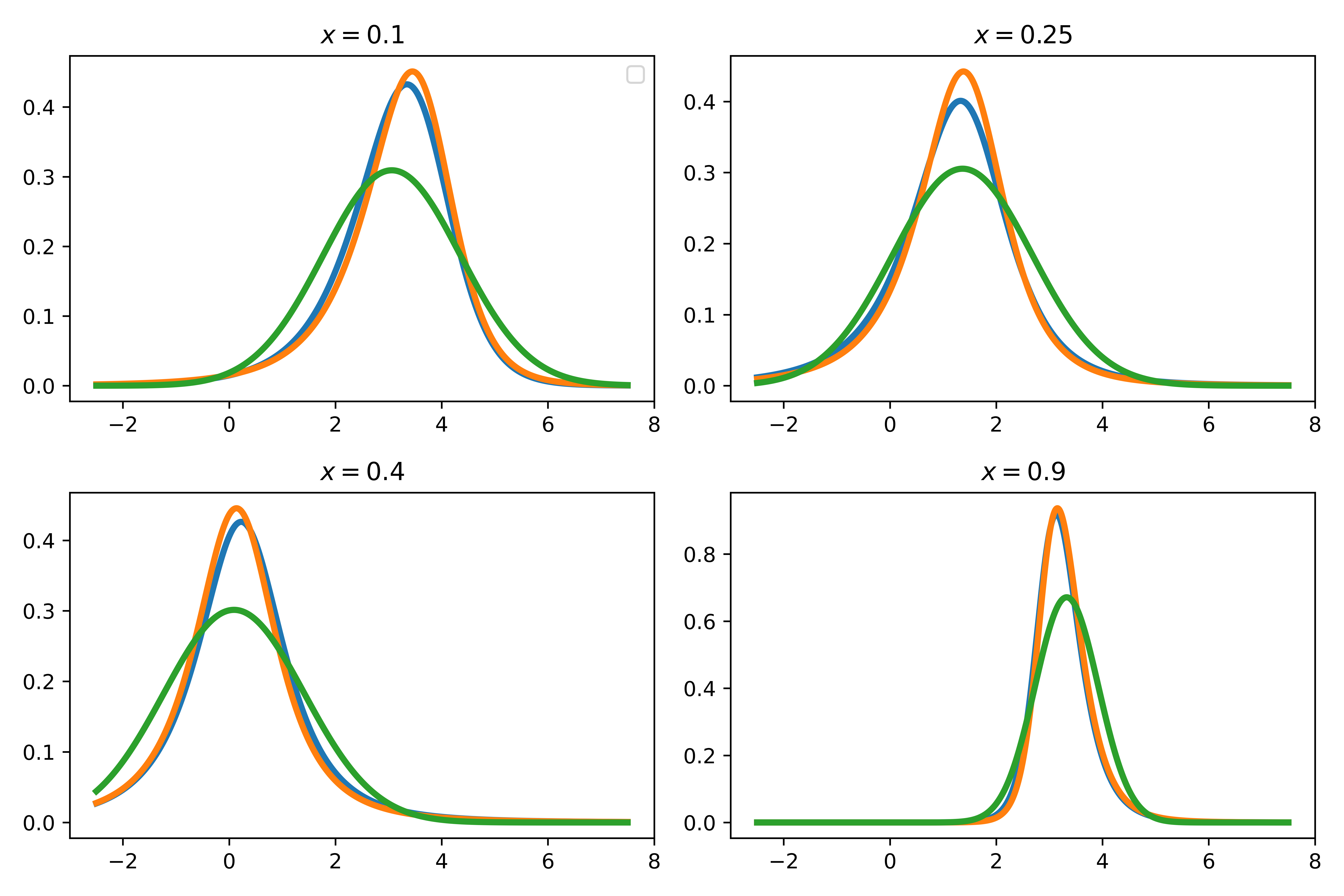}
    \caption{Comparison of true (blue) and fitted (Tukey
    g-and-h in orange vs Gaussian in green)
    probability density functions
    at four values of the scalar feature $x$,
    in the case where the target variable
    follows a Tukey g-and-h distribution.}
    \label{fig:fig2}
\end{figure}

In Figure~\ref{fig:fig1} we show the resulting fit between the
true regression functions and the functions learnt
by the neural network.
In Figure~\ref{fig:fig2} we compare the fitted Tukey-g-and-h
distribution, and a fitted Gaussian distribution for 
different values of the feature $X$. Unsurprisingly,
we observe that in the case of a left skew of the target
variable (e.g. x=0.1, top left), the mode of the fitted Gaussian distribution
is underestimated. On the contrary, in the case of a right
skew of the target variable (e.g. x=0.9, bottom right), the mode of the fitted
Gaussian distribution is overestimated.

\subsection{Model misspecification}
\begin{figure}
    \centering
    \includegraphics[width=0.48\textwidth]{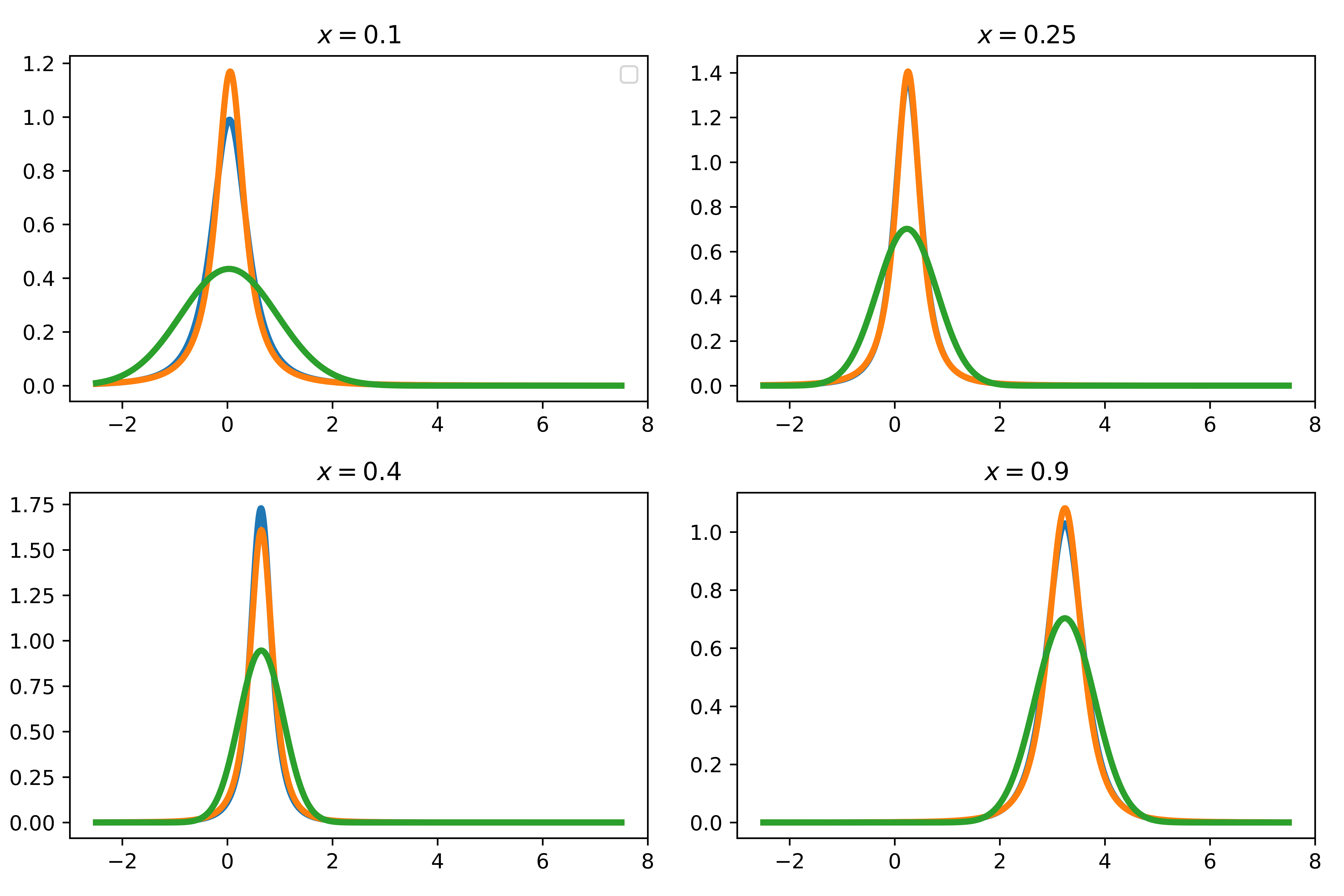}
    \caption{Comparison of true (blue) and fitted 
    (Tukey
    g-and-h in orange vs Gaussian in green)
    probability density functions 
    at four values of the scalar feature $x$,
    in the case where the target variable
    follows a t-distribution.}
    \label{fig:t-fit}
\end{figure}

\begin{figure}
    \centering
    \includegraphics[width=0.48\textwidth]{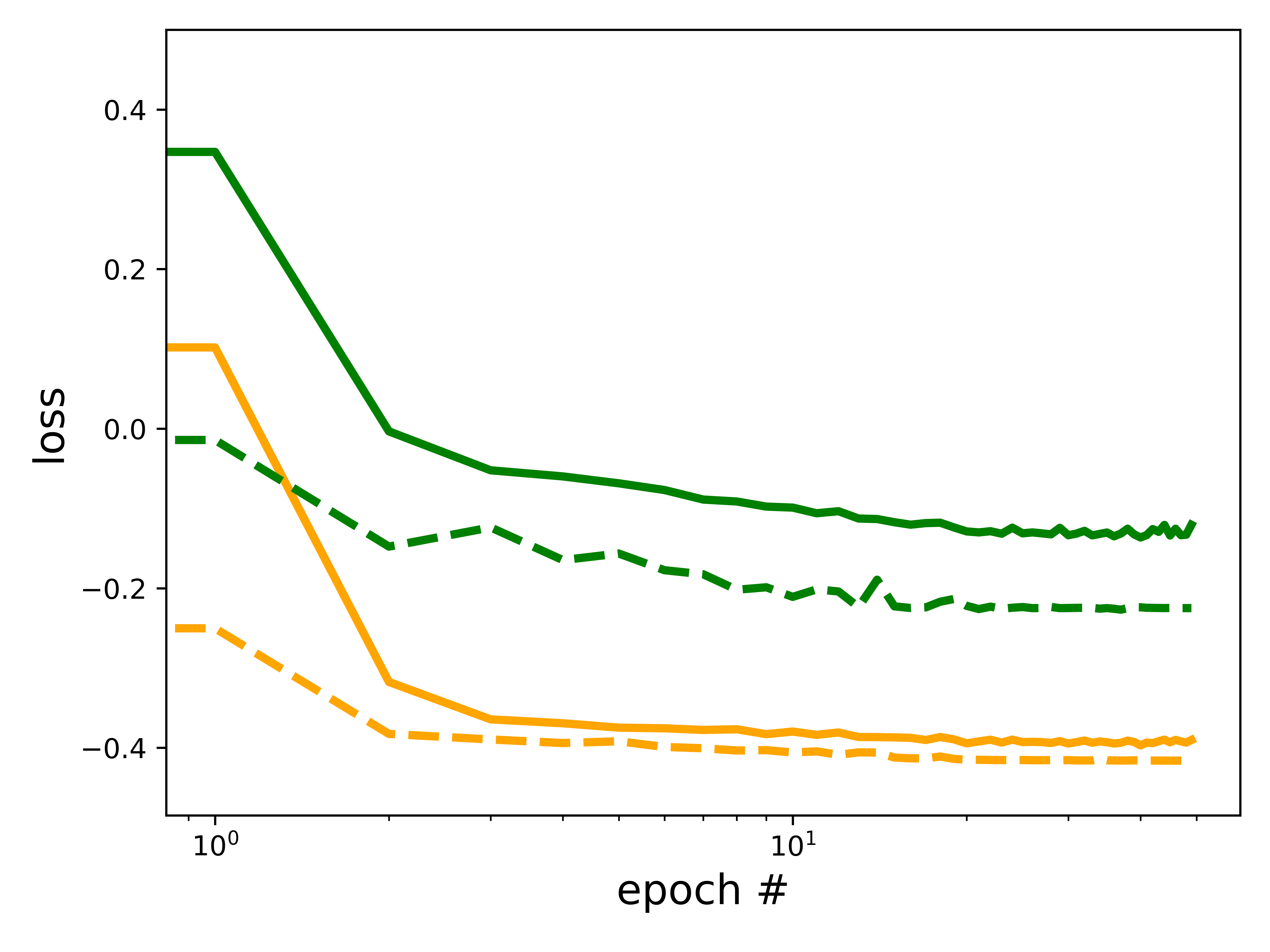}
    \caption{Tukey g-and-h (orange) and Gaussian (green) losses over
    training epochs for training (solid) and validation (dashed) data
    for a t-distributed target variable.}
    \label{fig:t-losses}
\end{figure}

We now present a simulated data experiment subject to model misspecification.
Specifically, we set the target
variable to follow a t-distribution,
the parameters of which are controlled
by arbitrarily specified functions,
\begin{equation}
    Y|X \sim \mu(X) + \sigma(X) T, \quad T\sim t_{\nu(X)},
\end{equation}
where $t_\nu$ is used to denote student's t-distribution with 
$\nu$ degrees of freedom.
Naturally, if we know the true distribution
of the target variable to follow a t-distribution, we are likely to be better off directly using that in our regression model. Here our intent is to show that our procedure is robust to a misspecified setting. 
In Figure~\ref{fig:t-fit} we compare the fitted distribution for 
four values of the scalar input. The Tukey g-and-h recovers the shape of
the distribution of the target variable, in comparison to the
Gaussian neural network regression. This is validated by the
training and validation losses, which are shown in Figure~\ref{fig:t-losses}. Finally, we carry out a
goodness-of-fit analysis by following
the procedure described in Section~\ref{sec:gof}, see
Figure~\ref{residuals_t} in Appendix~\ref{app:sim},
where we show a histogram of the $\widehat{z}_i$
residuals against the probability density function
of a standard normal variable.

\section{Real data experiment: an application to crop yield predictions}
Food security is widely recognised as one of the most urgent challenges we currently face globally \cite{lee2023ar6}. This concern has grown in significance due to the changing climate and its warming effects. With the increasing occurrence of extreme weather events and alterations in weather patterns, our food system, especially in certain regions, has become highly vulnerable to the impacts of climate change.  Agriculture is both one of the sectors most susceptible to climate change and a significant contributor to it \cite{mahowald2017impacts}. Therefore, it is essential to consider both mitigation and adaptation strategies, as well as transforming agricultural practices to promote sustainability and resilience. A key objective is to develop more reliable and scalable methods for monitoring global crop conditions promptly and transparently, while also exploring how we can adapt agriculture to mitigate the effects of climate change.

The key areas of research in AI for agriculture include crop mapping, crop type mapping, field boundary delineation, yield estimation, and pests and disease detection. While there are other use cases such as crop suitability, the aforementioned categories encompass the majority of research conducted in this field. We can determine yield at various scales, assess crop conditions, and combine the model outputs with other data sources, such as temperature and rainfall predictions, to derive comprehensive insights. These insights can aid in decision-making for farmers and policymakers, facilitating informed choices regarding yields, diseases, and other relevant factors. 

We propose to apply our proposed methodology to global crop yield prediction since food security in general and yield prediction in particular pose  one of the most pressing issues in AI for agriculture. Extreme events can lead to crop yield declines, resulting in financial losses and threats to food  security. Crop yield
prediction involves predicting the volume or weight of crops to be harvested in each unit area. This is a regression problem where the goal is to estimate crop yield per unit area, representing the rate of production. There is extensive literature on crop yield prediction, where it can be considered on a plant, farm, global or regional scales. In this work, we consider a global scale and therefore utilise a global dataset for crop yield prediction. Shuai et al.  predicted maize yield in the United States at the pixel level, providing estimates in tonnes per hectare \cite{SHUAI2022112938}. The paper by You et al. (2017) introduced a different approach, employing deep Gaussian processes to predict crop yield based on remote sensing data \cite{you2017deep}, with focus on estimating yield at the county scale rather than at the individual pixel level. 

\begin{figure}[h!]
    \centering
    \includegraphics[width=0.48\textwidth]{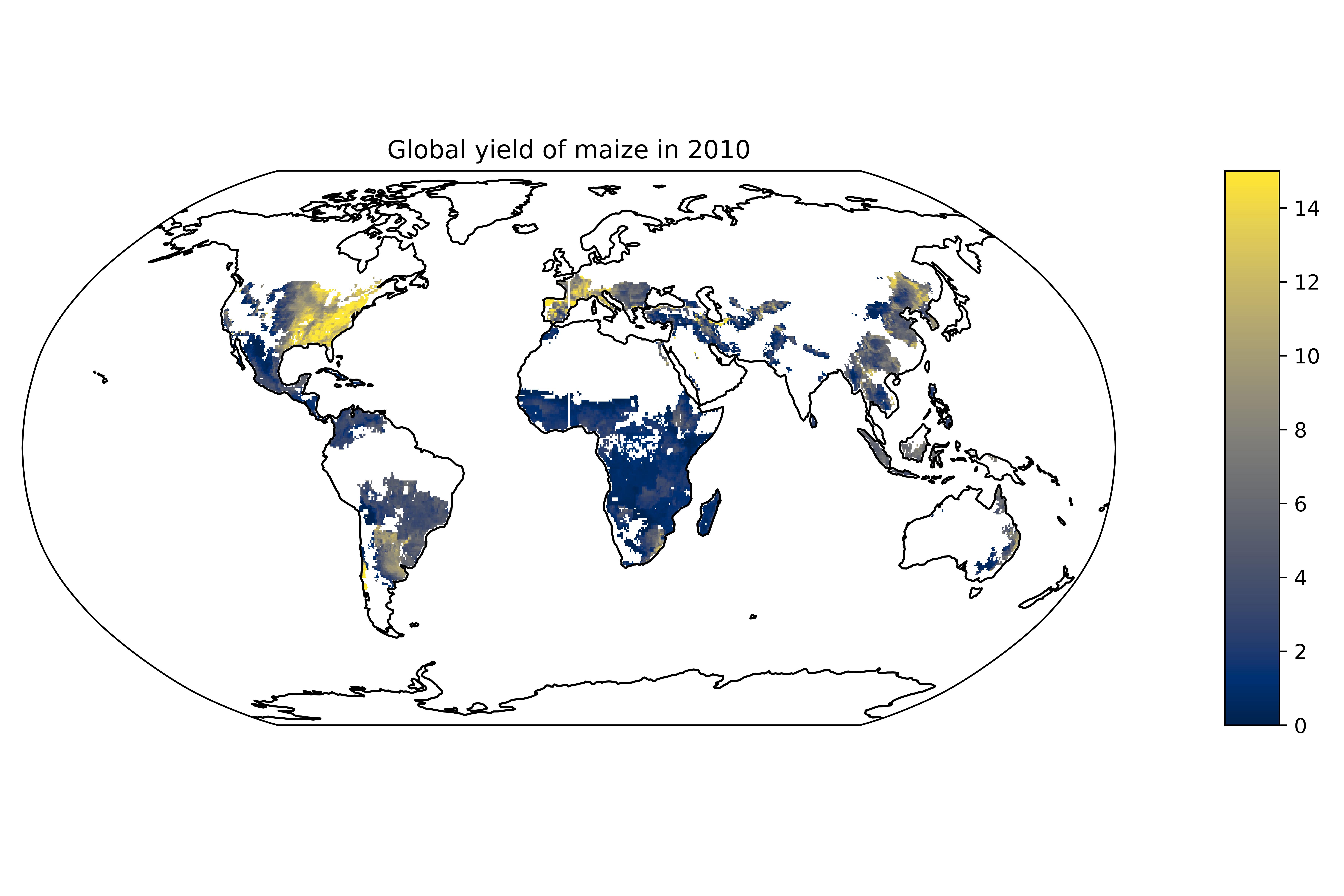}
    \caption{Global yield of maize
    in ton per hectare in 2010 on
    a 0.5' spatial-resolution grid}
    \label{fig:maize_2010}
\end{figure}

We investigate a yearly global yield dataset
\cite{iizumi2019gdoh}
for maize, wheat, rice, and soybean,
spanning the period 1981-2016.
For each year and each crop, the yield is provided
in tons per hectare
on a global grid with 0.5' spatial resolution. For instance, Figure~\ref{fig:maize_2010}
shows the global yields for maize in 2010.
Our goal is to learn, for each crop, a
parametric distribution of the
yield as a function of latitude, longitude, and year. 
We compare two approaches:
\begin{enumerate}
    \item Gaussian prediction, based on a negative Gaussian likelihood loss
    function.
    \item Tukey g-and-h prediction, based on a negative Tukey g-and-h likelihood
    loss function, as described in
    Section~\ref{sec:lkh}.
\end{enumerate}
The architecture of our neural network is a follows:
a sequence of fully connected layers with
ReLU activations and batch normalization.
The neural network takes the latitude and longitude
as inputs to the first hidden layer, and additionally
the year is concatenated to the penultimate
hidden layer.
The two approaches use the same neural
network architecture, with the exception that they
output a different number of values.
For the first approach, we output two values,
corresponding to the two parameters of
a Gaussian distribution, its mean and
variance.
For the second approach, we output four 
values, one for each of the four parameters
that define a Tukey g-and-h distribution.
Training is performed according to the
Adam algorithm~\cite{kingma2017adam}, and with the
use of a scheduler that reduces the
learning rate by a factor of 10 at epochs
10, 15, 20, 30 and 40 from an
initial value of $1\mathrm{e}{-4}$. Finally, we use
a batch size of 4096.

To compare the two approaches, we split the data, by using years 
1985, 1995, 2005, 2015 for validation 
and 1986, 1996, 2006, 2016 for testing. 
All remaining years from the dataset are used for training. 

\begin{figure}[h!]
    \subfloat[]{
        \includegraphics[width=0.48\textwidth]{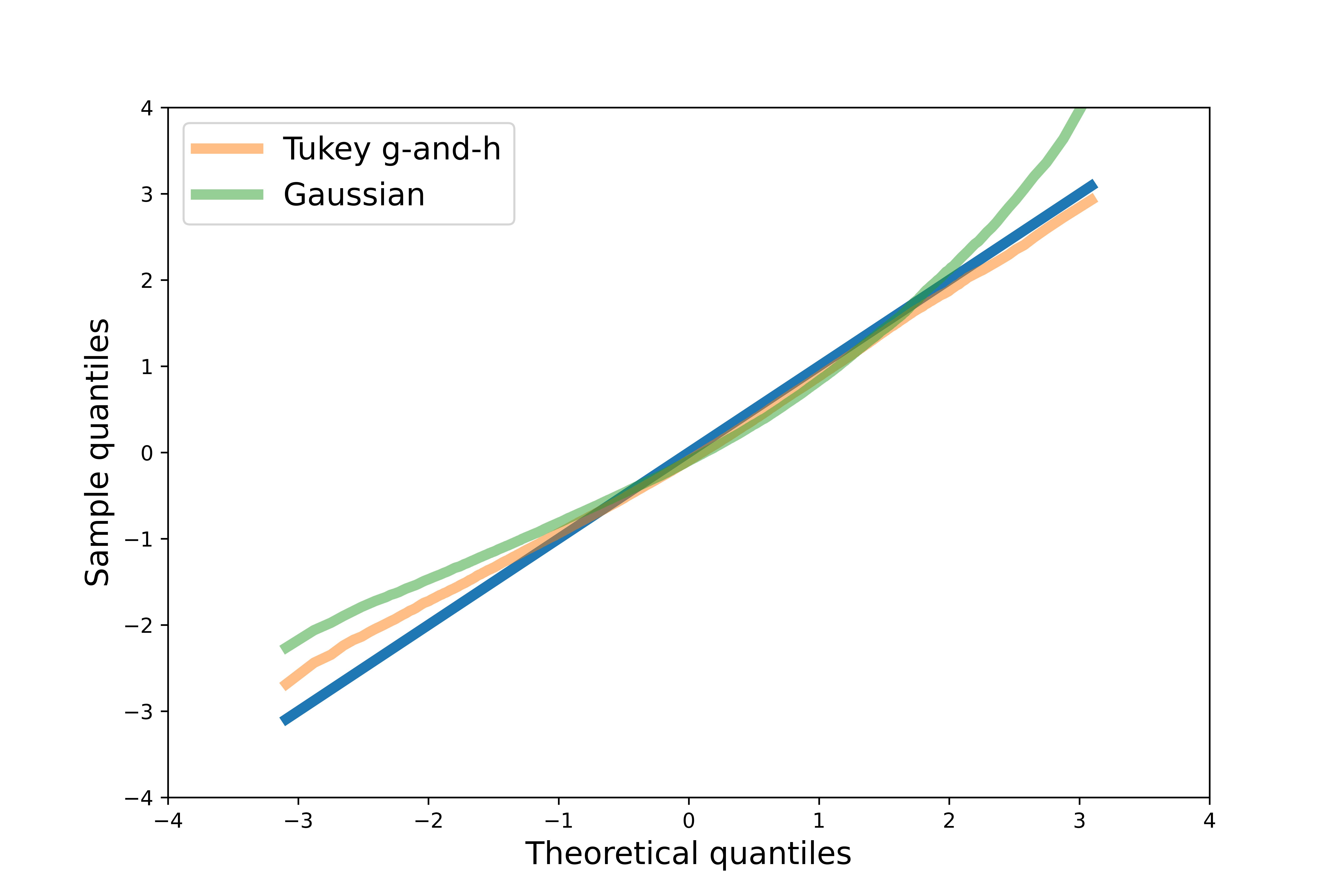}
        \label{fig:qq_val_crop}
    }\\
    \subfloat[]{
        \includegraphics[width=0.48\textwidth]{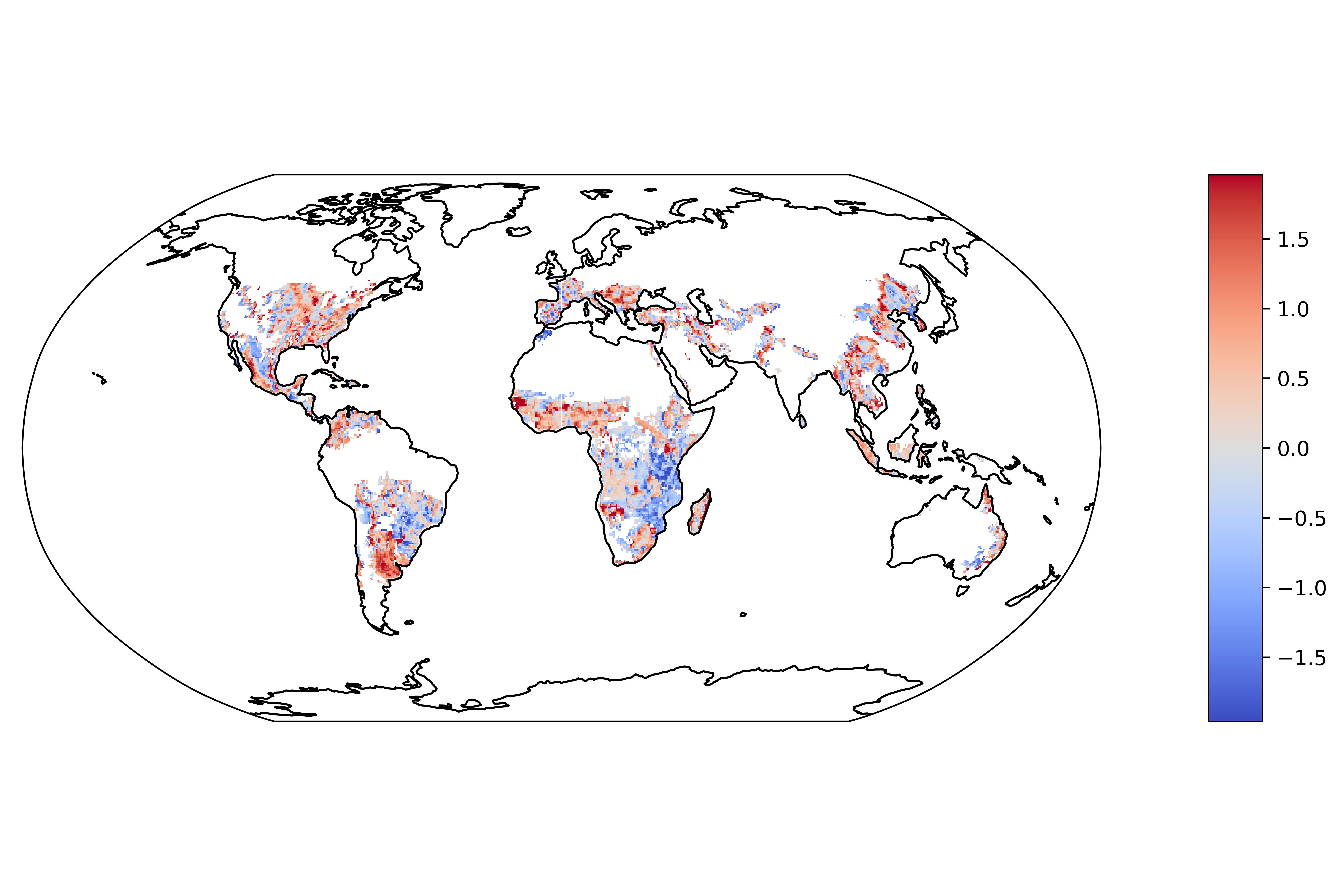}
        \label{fig:map_res_val_crop}
    }
    \caption{(a) quantile-quantile plot
    of the maize yield residuals from the validation
    data with respect to the standard
    normal distribution,
    (b) global map of residuals from the validation data (year 2005)}
    \label{fig:res_val}
\end{figure}

\begin{figure}[h!]
    \centering
    \includegraphics[width=0.4\textwidth]{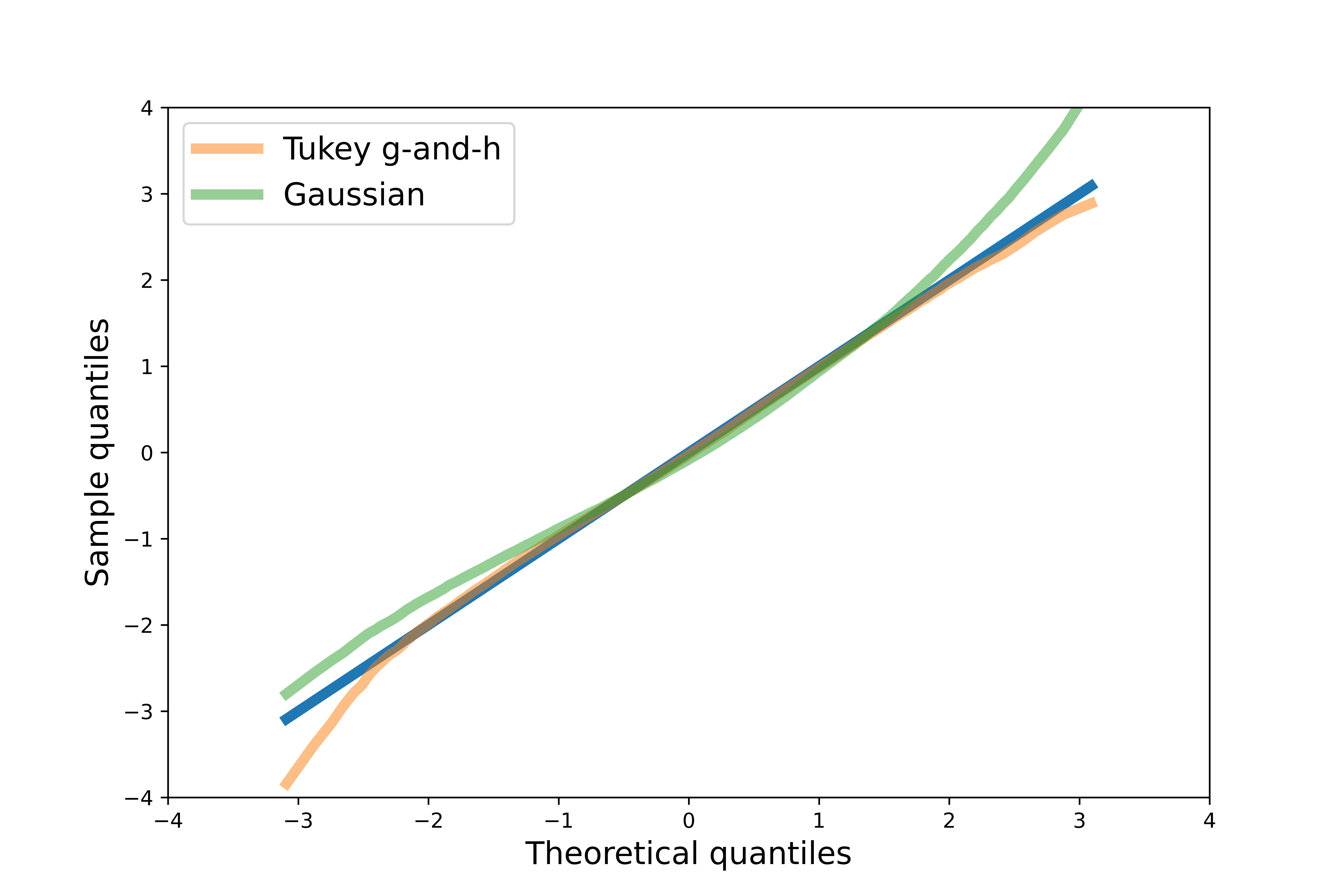}
    \caption{quantile-quantile plot
    of the maize yield residuals from the test data with respect to the 
    standard normal distribution}
    \label{fig:res_test}
\end{figure}

To assess each approach, we compute residuals
that are expected to follow a standard
normal distribution under the posited models,
according to the approach mentioned in Section~\ref{sec:gof}.
Therefore, we report quantile-quantile
plots of those residuals with respect
to the quantiles of the standard normal
distribution. This is shown
in Figure~\ref{fig:qq_val_crop} for the validation data
and Figure~\ref{fig:res_test} for the test data.
We note that for both the validation and test datasets,
the quantile-quantile plots show a better fit for the
proposed method that predicts a Tukey g-and-h distribution.

In future work, we will investigate the
use of weather indicators to more accurately
predict the yield. Average and extreme
weather conditions over the year naturally
play a key role in determining the yield.
Additionally, the spatial dependence
of weather conditions results in
spatial dependence of the yield, which
we wish to remove as much as possible
from a methodological point of view. 

\section{Conclusion}
The Tukey g-and-h distribution has a strong history of research
and applications in environmental sciences and other fields
such as financial modeling. It offers a good trade-off between
a limited number of parameters and the ability to approximate
a wide range of skewed and heavy-tailed probability
density functions. Applications of Deep Neural Networks
is also becoming more and more prevalent in the aforementioned 
fields. While the exact form of the Tukey g-and-h log-likelihood has no
known closed-form solution, we show in this paper that we can
still train a Deep Neural Network to predict a Tukey g-and-h
distribution. While the standard approach in
Tukey g-and-h random fields is to assume that
a unique Tukey g-and-h is applied pointwise,
here we allow the parameters of the transform
to be features-dependent, and learn the corresponding
mapping via standard neural network training techniques.
This can naturally be extended to multi-modal 
Tukey g-and-h
distributions with 4 output neurons for each mode and
a softmax over $p$ neurons where $p$ is the number of modes.
A natural direction
for future work would be to address the remaining
dependence in the target variable. For instance,
looking at Figure~\ref{fig:map_res_val_crop},
it is clear that there remains some spatial
dependence between residuals at neighbouring
locations. Additional features --- such as weather
patterns for this application--- may further 
account for this remaining dependence. However
for applications such as spatial interpolation,
it will be judicious to explicitly model and
estimate a Gaussian covariance kernel 
on the residuals.

\section*{Acknowledgments}
This research utilised Queen Mary's Apocrita HPC facility, supported by QMUL Research-IT. doi:10.5281/zenodo.438045. More specifically,
the OnDemand services~\cite{Hudak2018} have greatly facilitated the development
of the research outputs presented in this paper.

\newpage
{\appendices
\section{Additional figures
from the simulated data 
experiments}
\label{app:sim}
\begin{figure}[h!]
    \centering
    \includegraphics[width=0.48\textwidth]{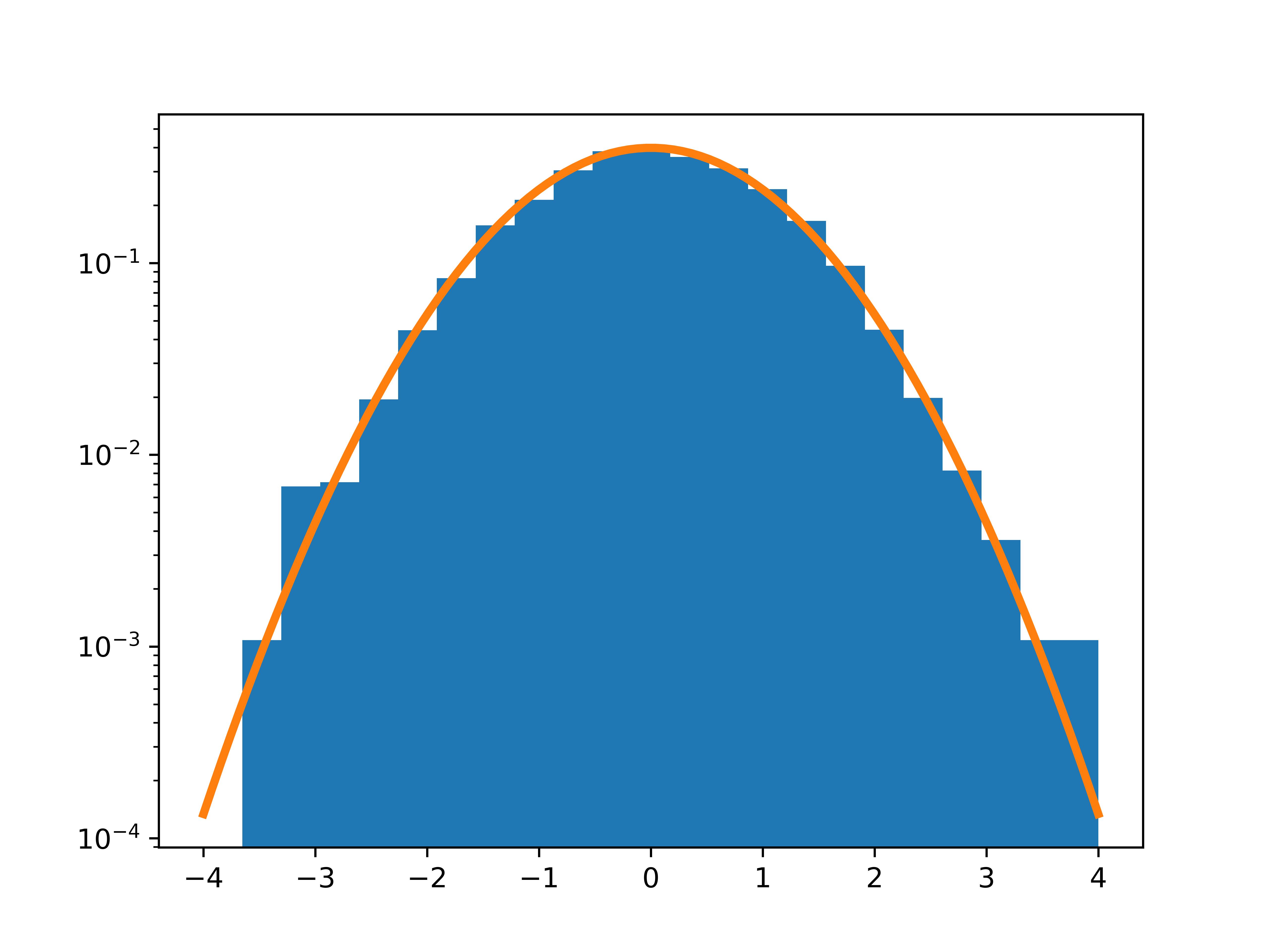}
    \caption{Log-density histogram of standardized residuals on validation data obtained from the Tukey g-and-h neural network trained on simulated t-distributed target variables versus the probability density function of a standard normal random variable.}
    \label{residuals_t}
\end{figure}

\section{Crop yield for maize}
\begin{figure}[h!]
    \centering\includegraphics[width=0.48\textwidth]{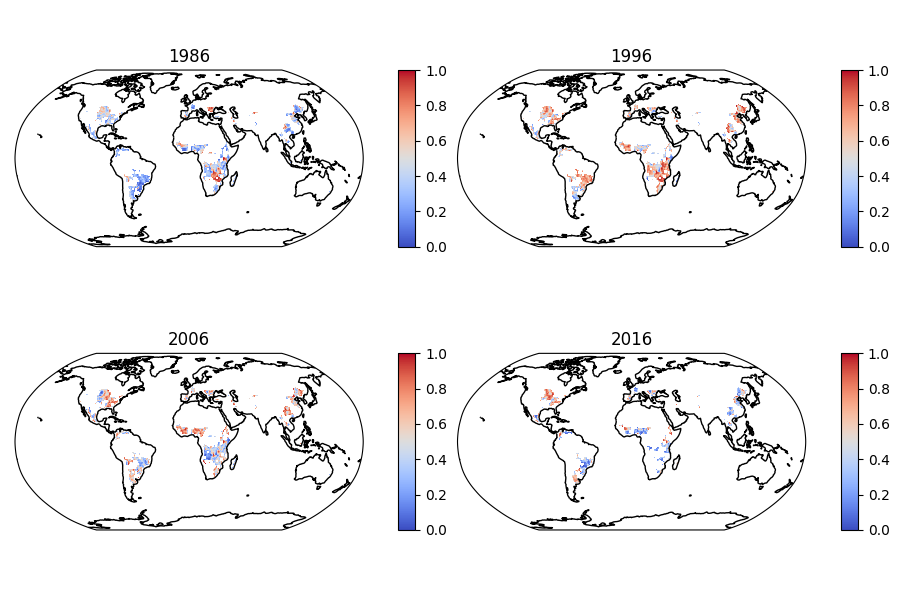}
    \caption{Map of uniform residuals
    for maize yield on test years}
\end{figure}
\newpage

\section{Crop yield for rice}
\begin{figure}[h!]
    \centering
    \includegraphics[width=0.48\textwidth]{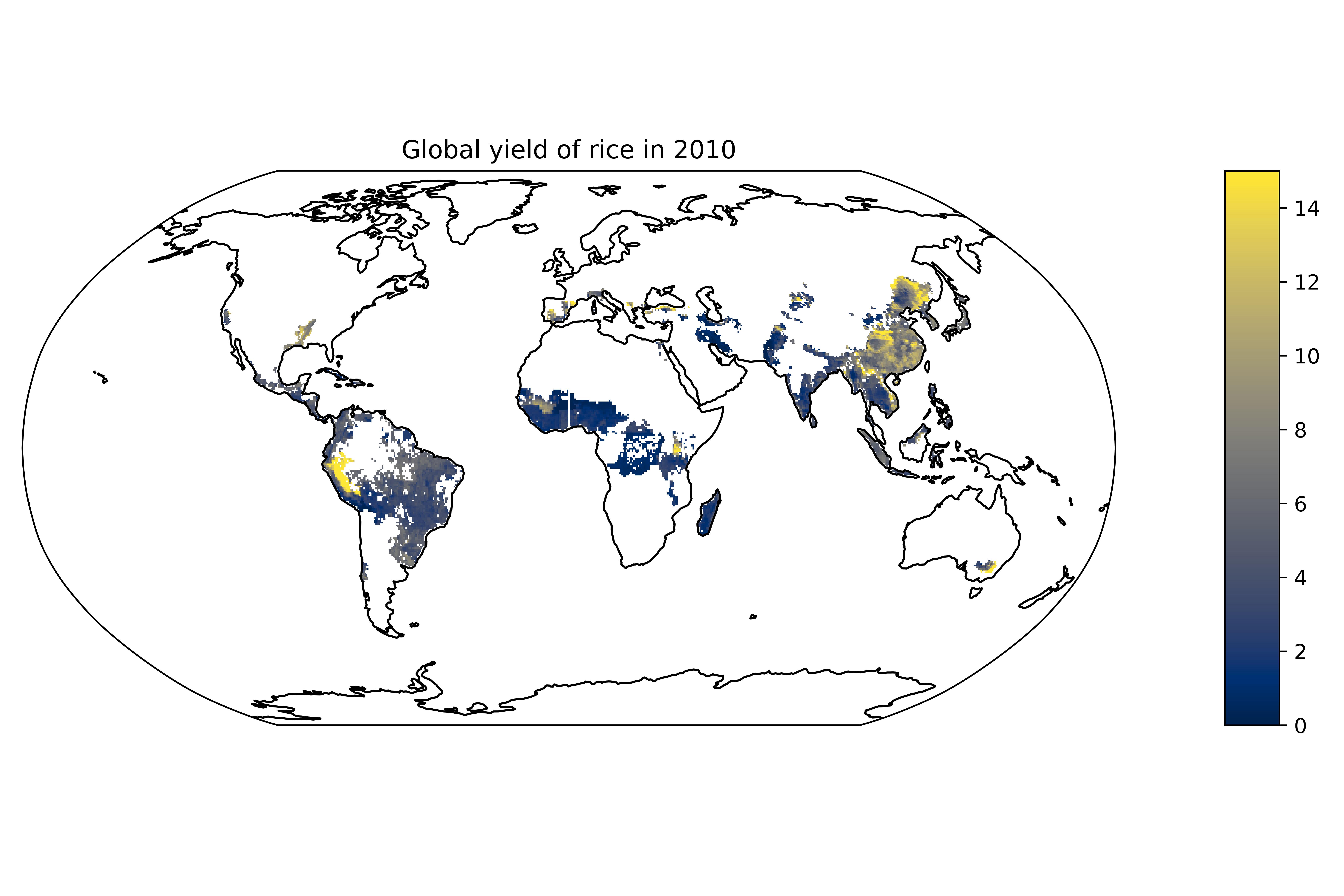}
    \caption{Gloabl yield of rice in ton per hectar in 
    2010 on a 0.5' spatial-resolution grid.}
\end{figure}
\begin{figure}[h!]
    \centering
    \includegraphics[width=0.48\textwidth]{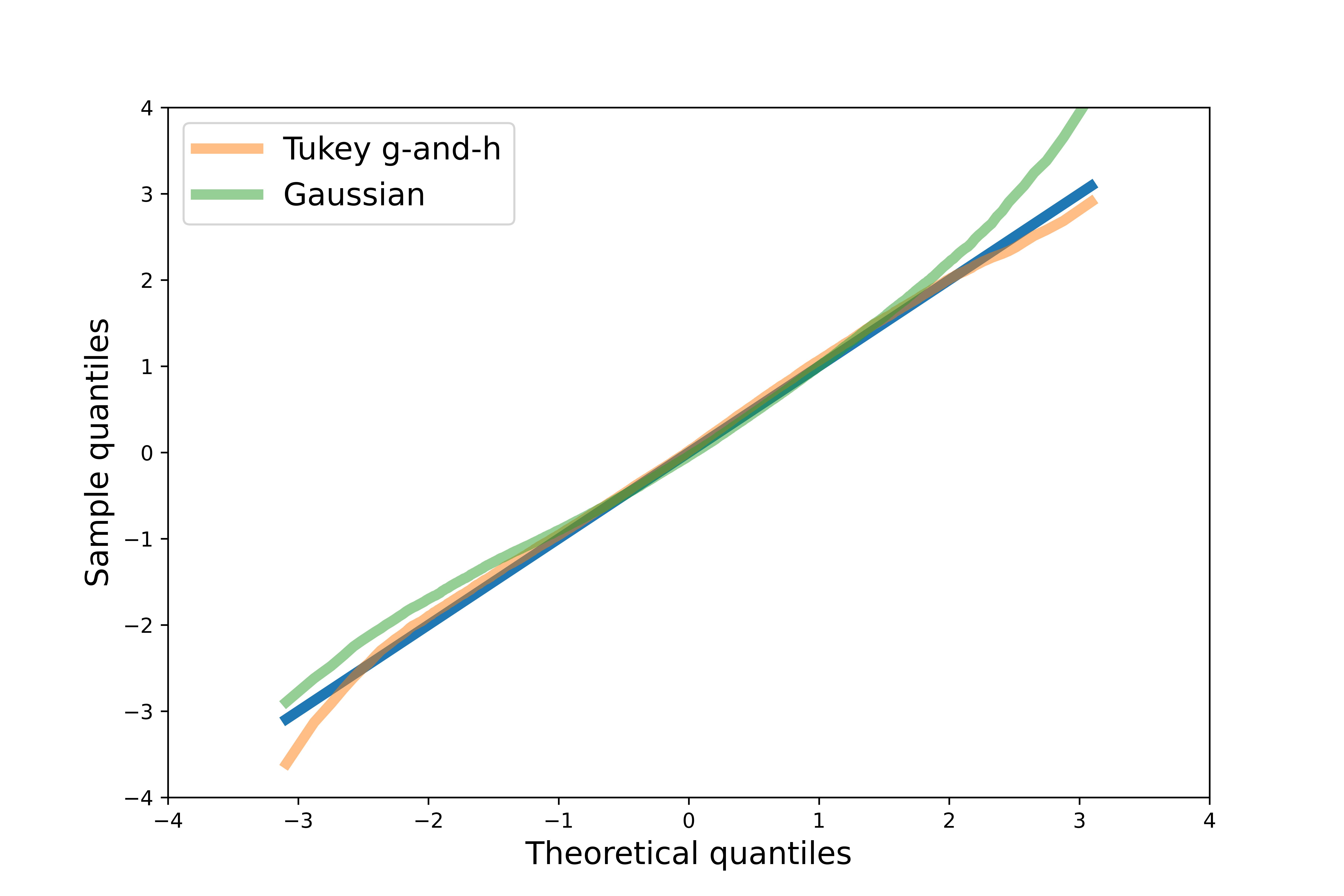}
    \caption{quantile-quantile plot of the rice residuals from
    the test data for methods 2 and 3 with respect to
    the standard normal distribution.}
\end{figure}

\newpage
\section{Crop yield for wheat}
\begin{figure}[h!]
    \centering
    \includegraphics[width=0.48\textwidth]{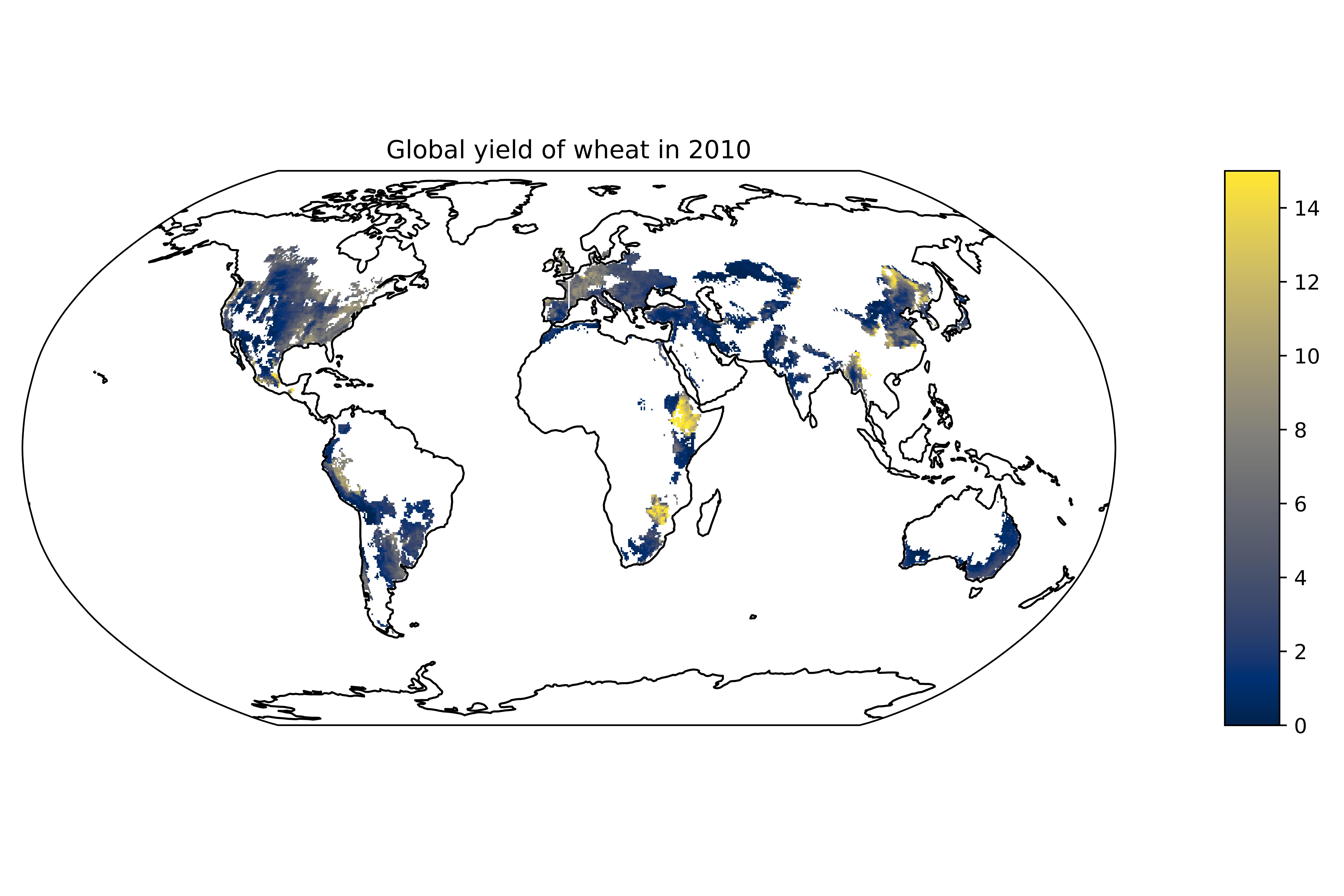}
    \caption{Gloabl yield of wheat in ton per hectar in 
    2010 on a 0.5' spatial-resolution grid.}
\end{figure}
\begin{figure}[h!]
    \centering
    \includegraphics[width=0.48\textwidth]{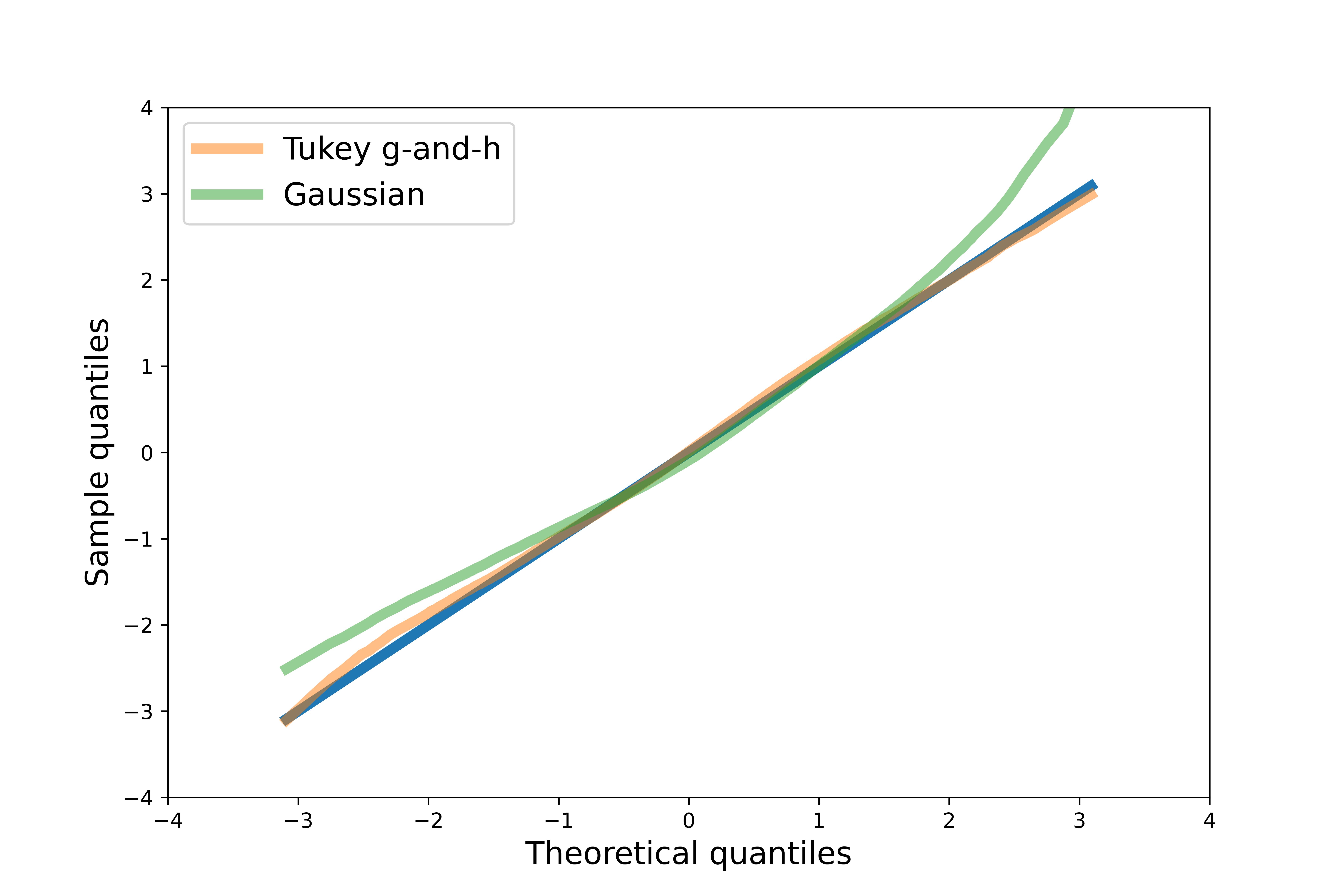}
    \caption{quantile-quantile plot of the wheat residuals from
    the test data for methods 2 and 3 with respect to
    the standard normal distribution.}
\end{figure}


\newpage
\bibliographystyle{ieeetr}
\bibliography{sample}

 
\vspace{11pt}

\vspace{11pt}

\vfill

\end{document}